\documentclass[11pt]{article}
\usepackage{emnlp2021}
\usepackage{times}
\usepackage{latexsym}

\usepackage{microtype}
\usepackage{amsmath}
\usepackage{arydshln}
\usepackage{multirow}
\usepackage{graphics}
\usepackage{url}
\usepackage{textcomp}
\usepackage{mathtools}
\usepackage{amsfonts}

\title{Attention-based Contrastive Learning for Winograd Schemas}

\author{Tassilo Klein \\
  SAP AI Research \\
  \texttt{tassilo.klein@sap.com} \\\And
  Moin Nabi \\
    SAP AI Research \\
  \texttt{m.nabi@sap.com} \\}

\date{}

\begin{document}
\maketitle
\begin{abstract}
Self-supervised learning has recently attracted considerable attention in the NLP community for its ability to learn discriminative features using a contrastive objective~\cite{qu2020coda,klein-nabi-2020-contrastive}. This paper investigates whether contrastive learning can be extended to Transfomer attention to tackling the Winograd Schema Challenge. To this end, we propose a novel self-supervised framework, leveraging a \emph{contrastive loss} directly at the level of \emph{self-attention}. Experimental analysis of our attention-based models on multiple datasets demonstrates superior commonsense reasoning capabilities. The proposed approach outperforms all comparable unsupervised approaches while occasionally surpassing supervised ones.\footnote{The source code can be found at: \url{https://github.com/SAP-samples/emnlp2021-attention-contrastive-learning/}} \end{abstract}

\section{Introduction}

Pre-trained language models have propelled the domain of NLP to a new era. Specifically, Transformer-based models are the driving force behind recent breakthroughs. However, despite all the recent success in text understanding, the task of commonsense reasoning is still far from being solved~\cite{marcus2020next,kocijan2020review}.
In order to assess the commonsense reasoning capabilities of automatic systems, several tasks have been devised. Among them is the popular Winograd Schema Challenge (WSC)~\cite{levesque2012winograd}.
WSC frames commonsense reasoning as a pronoun co-reference resolution problem \cite{lee2017end}, which consists of twin-pair sentences. Experts curated the twin pairs manually to be ``Google-proof'', e.g., simple statistical biases from large data should be insufficient to resolve the pronouns. Hence, solving WSC was expected to require diverse reasoning capabilities (e.g., relational, causal).
Sentences in the twin pairs differ only in the ``trigger word''. Furthermore, trigger words are responsible for switching the correct answer choice between the questions.
Below is a popular example from WSC. In the example, the trigger word is underlined. The challenge entails resolving the pronoun ``it'' with a noun from the candidate set (``suitcase'', ``trophy''):
 \\
\\
\emph{\textbf{Sentence-1:}} \emph{The trophy doesn\textquotesingle t fit in the suitcase because {\textbf{it}} is too \underline{small}.}\\
\emph{\textbf{Answers:}} \emph{\textbf{A)}} the trophy \emph{\textbf{B)}} the suitcase\\ \\
\emph{\textbf{Sentence-2:}} \emph{The trophy doesn\textquotesingle t fit in the suitcase because {\textbf{it}} is too \underline{big}.}\\
\emph{\textbf{Answers:}} \emph{\textbf{A)}} the trophy \emph{\textbf{B)}} the suitcase
\\
\\
The research community has recently experienced an abundance of methods proposing to utilize the latest language model (LM)
for commonsense reasoning~\cite{kocijan19acl, he2019hybrid, ye2019align, ruan2019exploring, trinh2018simple, klein-nabi-2019-attention, tamborrino2020pre}. Models learned on large text corpora were hoped to internalize commonsense knowledge implicitly encountered during training.
Most of such methods approach commonsense reasoning in a two-stage learning pipeline. Starting from an initial self-supervised learned model, commonsense enhanced LMs are obtained in a subsequent fine-tuning (ft) phase. Fine-tuning enforces the LM to solve the downstream WSC task only as a plain co-reference resolution task.
Despite some initial success in this direction, we hypothesize that the current self-supervised tasks used in the pre-training phase are too ``shallow'' to enforce the model to capture a ``deeper'' notion of commonsense~\cite{kejriwal2020finetuned,winograd-elazar}.
Shortcomings of models obtained in such a fashion can partially be attributed to the training corpora itself. Standard training sets such as Wikipedia barely contain commonsense knowledge,
so supervised fine-tuning only promotes the discovery of ``artificial'' cues and language biases to tackle commonsense reasoning~\cite{trichelair2018evaluation,DBLP:journals/corr/abs-1810-00521,trichelair-etal-2019-reasonable,emami2019knowref, kavumba2019choosing}.
This is the main reason why supervised methods pre-trained on large datasets (e.g., WinoGrande) can transfer effectively to smaller target datasets (e.g., WSC) yet do not show the same performance level on the source dataset.
\\
In an attempt to avoid the utilization of shallow commonsense reasoning cues, very recently~\cite{klein-nabi-2020-contrastive} introduced a Contrastive Self-Supervised (CSS) learning method, leveraging the mutual-exclusivity of WSC pairs. Despite almost reaching state-of-the-art performance, the approach does not require external knowledge for training.  However, the authors observed that leveraging the contrastive loss directly on the Transformer-backbone at the \emph{LM-level} can destabilize the self-supervised optimization.
\\
We propose a novel self-supervised loss to address this, introducing an abstraction layer between the backbone and the downstream task.
Our approach smoothly manipulates the attentions to achieve this goal in a Transformer-like fashion while avoiding destabilization of the intrinsics. To do so, we make use of the non-identifiability property of attention, which implies that the attention values are not uniquely determined from the head’s output, and vice versa. Consequently, various attention patterns across the Transformer can result in identical outcomes and permit regularization - see for  details~\cite{Brunner2020On}.
Intuitively, the proposed contrastive attention mechanism does not overwrite the low-level semantics captured in the pre-trained model. Instead, it induces modest adjustments via attention patterns.
In the context of Winograd schemas, the proposed approach shifts the attention from the wrong answer candidate to the right candidate. Simultaneously, the attention contrast forces the LM to be more rigorous across attention heads while consistent over the samples.
\\
In summary, our contributions are the following:
\textbf{First}, we propose a contrastive loss enforced on the Transformer attention, which helps for the emergence of commonsense patterns. \textbf{Second}, we present empirical evidence showcasing the viability of the approach, outperforming comparable state-of-the-art.

\section{Attention-based Contrastive Learning}
\begin{figure*}
    \centering
    \includegraphics[width=0.85\textwidth]{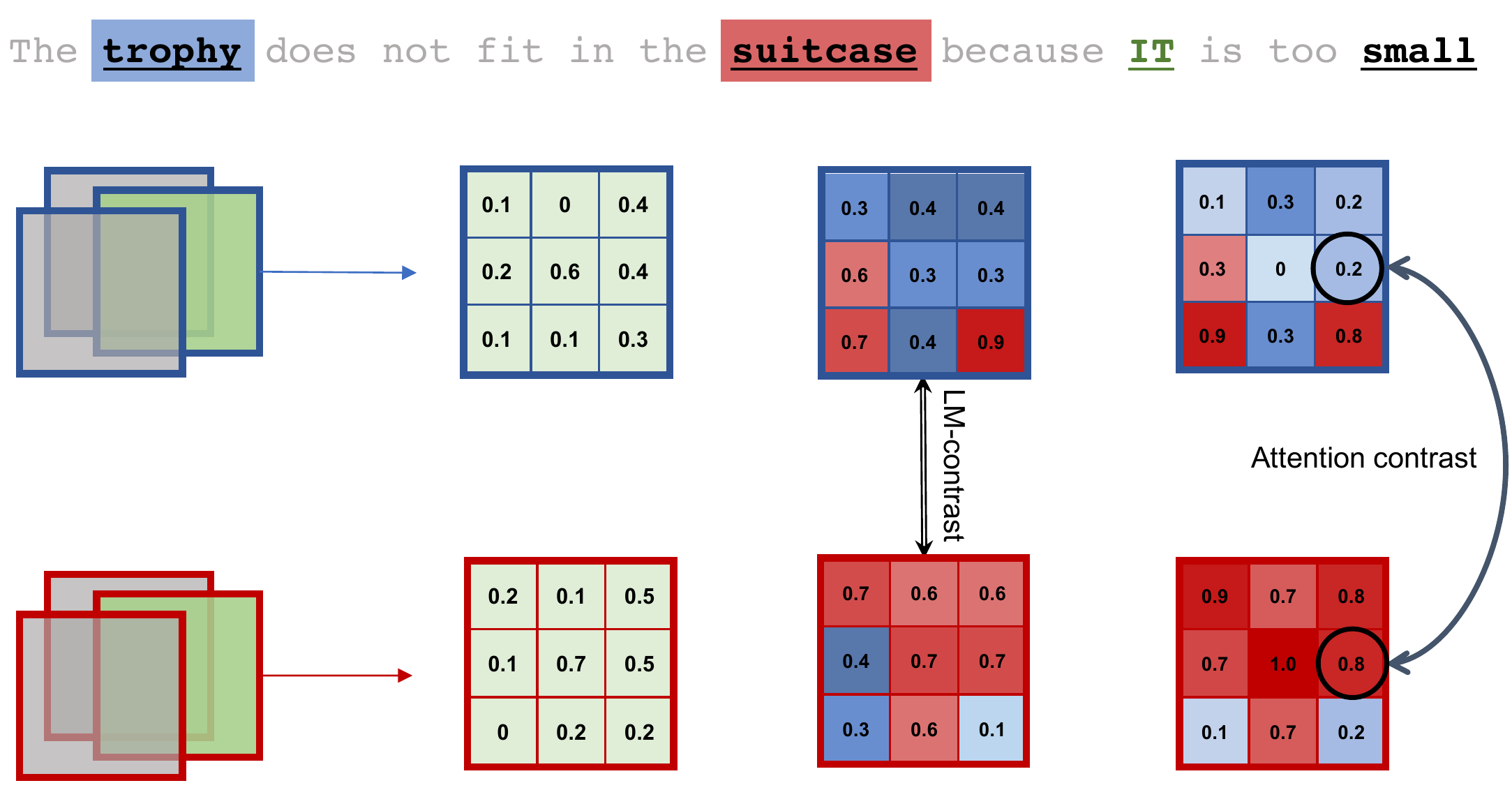}
    \caption{Schematic illustration of contrastive learning for a particular sentence, where colors show attention maps for different words of a mock setup with $3$ heads and $3$ layers. Squares with blue/red frames correspond to specific sliced attention $3\times3$ matrix  for candidates, establishing the relationship to the reference pronoun indicated with green. Attention is color-coded in blue/red for candidates “trophy”/ “suitcase”; the associated pronoun “it” is indicated in green. The \textbf{Attention-based Contrast} shows a more consistent disambiguation attention for the correct candidate compare to the \textbf{LM-based Contrast}~\citep{klein-nabi-2020-contrastive}. }
    \label{fig:main}
\end{figure*}
\noindent\textbf{Preliminaries: }The proposed approach extends the contrastive self-supervised method~\cite{klein-nabi-2020-contrastive} to facilitate commonsense reasoning for Winograd schemas at the attention level. In the context of data, we assume that $\mathcal{D}$ with $N=|\mathcal{D}_c|$ is a dataset constructed from contrastive twin-pairs samples, $\left(s_i, s_{i+1}\right) \in \mathcal{D}_c$, with $c_j$ and $c_{j+1}$ denoting answer candidates. The difference between the sentence pairs is the so-called ``trigger words'' responsible for flipping the answer in pronoun disambiguation. Thus, this trigger-word structure induces a mutual-exclusive candidate answer relationship at the pair level.
In the context of the model, we employ a Transformer-based LM for Masked Token Prediction~\cite{devlin2018bert}. Given a sentence with a {\tt [MASK]} token, the LM provides the likelihood of sentence $s_i$ with the token replaced by candidate tokens  $c_j \in \{\texttt {[CANDIDATE-1], [CANDIDATE-2]}\}$ denoted as   $p\left(c_j|s_i\right)=p_{i,j}$, assuming that the dataset consists of $i\in\frac{N}{2}$ distinct twin-pairs. Besides sentence likelihoods, the Transformer architecture also provides an attention tensor $\mathcal{A}(x) \in \mathbb{R}^{H\times L \times\mathcal{C}\times\mathcal{C}}$, for a given an input $x$ with $|x| = \mathcal{C}$, where $L$ denotes the number of layers, and $H$ the number of heads. Then the tensor decomposes into elements ${a}^{h,l}_{i,j}(x)$, gauging the influence of token $i$ w.r.t. token $j$ in layer $l$ of attention head $h$.

\subsection{Method}
Inspired by \citep{klein-nabi-2020-contrastive}, we make use of the structural prior of Winograd schemas and their within-pair mutual-exclusivity. We formulate this as in context of Transformer-based LM as a multi-task optimization problem defined as:
\begin{equation*}
\mathcal{L}(f_\theta) = \mathcal{L}(f_\theta)_{CM} + \mathcal{L}(f_\theta)_{CM}
\end{equation*}
Here $f$ denotes the underlying LM parameterized by $\theta$.
The first term, $\mathcal{L}_{CM}$ leverages the contrast arising from twin pairs enforcing mutual-exclusivity on attentions. The second term,  $\mathcal{L}_{CM}$, seeks to further reduce ambiguity at the LM level by maximization between the differences of the likelihoods for the answer candidates. It should be noted that although the proposed approach leverages the structural prior of twin pairs and it does not make use of any class label information explicitly, similar to~\cite{klein-nabi-2020-contrastive}.
See Fig.~\ref{fig:main} for a schematic illustration of the proposed method.
\subsubsection{Contrastive Attention }
\label{sec:CA}
The contrastive mechanism targets regularizing self-attention patterns emerging by invoking the LM on an input sequence, thus providing the model with commonsense reasoning capabilities. Specifically, the proposed approach seeks to induce \emph{consistently} higher attention values across all attention heads and layers for the right candidate as opposed to the wrong one. This contrasts with the LM-level MEx~\cite{klein-nabi-2020-contrastive}, where only the overall value of attention is enforced to be higher for the right candidate - see Fig. \ref{fig:main} for an illustration.
Hence, the proposed approach promotes the emergence of diverse attention patterns between the attention heads, avoiding issues such as the collapse to a single dominant head.
To this end, our proposed approach invokes twin-pair contrast on \emph{attention level} for samples in $\mathcal{D}_c$. This pushes for the superior establishment of distant dependencies more indirectly than enforcing it directly on the LM. Given the observation of~\cite{Brunner2020On} that distant relationships are formed towards the end of the transformer stack, we restrict instantiation of the contrastive attention loss on the last layers. This, in combination with the non-uniqueness of Transformer attentions w.r.t. output, operating on attention level suggests comparably smoother behavior. In order to resolve ambiguity in the attention mechanism w.r.t. candidates, we tie mutual exclusivity together with a binarization scheme. Here binarization refers to a simple form of mutual exclusivity loss applied in binary classification cases (such as WSC), defined as:
\vspace{-2mm}
\begin{multline*}
    \mathcal{L}_{CA}=-\lambda\sum_{ i=1,j=1\atop i\mathrel{+{=}}2 }^{N,2}\left(\mathbf{a}_{i,j}-\frac{\mathbf{e}}{2}\right)^2+\left(\mathbf{a}_{i+1,j}-\frac{\mathbf{e}}{2}\right)^2 \\
    +1- (\mathbf{a}_{i,j}-\mathbf{a}_{i+1,j})^2+1-[(1-\mathbf{a}_{i,j})-(1-\mathbf{a}_{i+1,j})]^2
\end{multline*}
Here $\mathbf{a}\in \mathbb{R}^H$ denotes a vector containing the attentions of all heads. Assuming attentions to be $\emph{normalized}$  w.r.t. candidates, i.e., $\sum_j a_{i,j} = 1$, effectively turns them into pseudo-likelihoods. Furthermore, $\mathbf{e}\in \mathbb{R}^H$ is vector with all elements 1, and $\lambda\in\mathbb{R}$ a hyperparameter.

\begin{table*}[]
\label{tab:my-table}
\begin{tabular}{l|ccccc}
 \textbf{Method} & \textbf{WSC} & \textbf{DPR} & \textbf{W.G.} & \textbf{K.Ref} & \textbf{W.Gen.} \\
 \hline
 \hline
Bi-LSTM~\cite{opitz2018addressing} & 56.0 & 63.0 & - & - & -\\
\hdashline
BERT (\texttt{DPR-ft})& 69.8 & - & 50.2 & {61.0} & 59.2\\
BERT (\texttt{MaskedWiki-DPR-ft})~\cite{kocijan19acl}
& 67.0 & 83.3 & 50.2 & - & \textbf{79.2}\\
BERT (\texttt{WikiCREM-DPR-ft})~\cite{kocijan19emnlp}  & 71.8 & 84.8 & - & - & -\\
\hdashline
RoBERTa (\texttt{DPR-ft})& 83.1 & - & \textbf{59.4} & 84.2 & -\\
RoBERTa (\texttt{WG-ft})~\cite{WinoGrande}
& \textbf{90.1} & \textbf{92.5} & - & \textbf{85.6} & -\\
\hline\hline
\cite{rahman-ng-2012-resolving} & 58.0 & 73.0 & - & - & - \\
\cite{peng-etal-2015-solving} & - & {76.4} & - & - & - \\
Knowledge Hunter~\cite{emami2018knowledge} & 57.1 & - & - & - & - \\
E2E~\cite{emami2019knowref}  & - & - & - & {58.0} & - \\
MAS~\cite{klein-nabi-2019-attention}
& 60.3 & - & - & - & - \\
Ensemble LM~\cite{trinh2018simple} & {63.8} & - & - & -& - \\
\hdashline
BERT (\texttt{zero-shot})~\cite{vaswani2017attention} & 62.6 & 58.5 & 51.7 & 62.3 & 62.5 \\
RoBERTa (\texttt{zero-shot})~\cite{liu2019roberta} & 67.7 & 70.3 & 53.7 & 60.4 & 61.6 \\
Self-supervised Ref. (\texttt{BERT})~\cite{klein-nabi-2021-towards} & 61.5  & 61.3 & 52.3 & 62.4 & 62.0 \\
Self-supervised Ref. (\texttt{RoBERTa})~\cite{klein-nabi-2021-towards} & 71.7 & 76.9 & 55.0 & 63.9 & 69.1 \\
\hdashline
CSS (\texttt{BERT})~\cite{klein-nabi-2020-contrastive} & 69.6 & 80.1 & 50.9 & 65.5& 69.5 \\
CSS (\texttt{RoBERTa})~\cite{klein-nabi-2020-contrastive} & 79.8 & \textbf{90.6} & 57.7 & 68.0 & 76.2 \\
\hline
\textbf{Our Proposed Method} & \textbf{84.1} & {90.0} & \textbf{60.8} & \textbf{69.9} & \textbf{93.3}\\
\hline
\end{tabular}
\caption{Results on different tasks: WSC, DPR, WinoGrande(W.G.), KnowRef (K.Ref) and WinoGender (W.Gen). Task performances in accuracy (\%) are subdivided into two parts. Top: supervised (ft), bottom: unsupervised.
}
\label{tab:results}
\end{table*}

\subsubsection{Contrastive Margin}
\label{sec:CM}
To stabilize optimization, we leverage consistency between sentences of each contrastive pair. On the one hand, it leads to faster convergence. On the other hand, it enforces smoothness on the loss surface and decreases the overall gradient fluctuation. The CM term seeks to maximize the margin between the LM likelihoods for each candidate in a pair:
\begin{equation*}
     \mathcal{L}_{CM}=-\alpha\sum_{i,j}^{N,2} \max\left(0,|p_{i,j}-p_{i,j+1}|+\beta\right),
\end{equation*}
with $\alpha, \beta \in \mathbb{R}$ being hyperparameters.
\\
When training the language model, the algorithm will look for a pattern of consistency in the attention heads and layers rather than force-fit supervisory signals from labels.  Assuming the answer of the first sentence is {\tt [CANDIDATE-1]}, it follows the answer for the second one is {\tt [CANDIDATE-2]}. This restricts the answer space. As the model is forced to leverage the pairwise relationship to resolve the ambiguity, it needs to generalize w.r.t. commonsense relationships. Intuitively speaking, as no labels are provided to the model during training, the model seeks to make the answer probabilities less ambiguous. It should be noted that the proposed approach leverages the structural prior of twin pairs, not making use of any label.

\section{Experiments and Results}
\subsection{Setup}
We leverage RoBERTa~\cite{liu2019roberta} as Language Model for Masked Token Prediction, and DPR~\cite{rahman-ng-2012-resolving} as dataset for training. Specifically, we use the Hugging Face~\cite{Wolf2019HuggingFacesTS} implementation of RoBERTa. The model is trained for $22$ epochs using a batch size of $18$ (pairs). Hyperparameters are $\alpha=0.05$, $\beta=0.02$, $\lambda=1.0$. For optimization Adam was selected with a learning rate of $10^{-5}$. Commonsense reasoning is approached by first fine-tuning the pre-trained RoBERTa (\emph{large})  masked-LM model on the DPR~\cite{rahman-ng-2012-resolving}.
\subsection{Results}
While observing loss fluctuations by learning mutual-exclusivity at LM model directly via log-likelihood (MEx)~\cite{klein-nabi-2020-contrastive}, such fluctuations are less pronounced when operating at attention level (proposed approach).

We evaluate the performance on different tasks - see Tab.~\ref{tab:results}. As can be seen, the proposed approach outperforms other unsupervised methods by a significant margin, outperforming some supervised methods or at least significantly reducing the gap between supervised and unsupervised approaches. The results are discussed separately for each benchmark below:
\\
\noindent\textbf{\emph{WSC}}~\cite{levesque2012winograd}: the most well-known pronoun disambiguation benchmark. Our method outperforms the strongest unsupervised baseline CSS(\texttt{BERT}) margin of $(+14.5\%)$ and CSS(\texttt{RoBERTa}) by $(+4.3\%)$.
\\
\noindent\textbf{\emph{DPR}}~\cite{rahman-ng-2012-resolving}: this pronoun disambiguation benchmark resembles WSC, yet significantly larger in size. According to ~\cite{trichelair2018evaluation}, less challenging due to inherent biases. Here the proposed approach outperforms the unsupervised baseline CSS(\texttt{BERT}) by a margin of $(+9.9\%)$, while observing a slight drop of $(-0.6\%)$ compared to CSS(\texttt{RoBERTa}).
\\
\noindent\textbf{\emph{WinoGrande (W.G.)}}~\cite{WinoGrande}: the largest dataset for Winograd co-reference resolution. Our method outperforms the unsupervised baseline CSS(\texttt{BERT}) by $(+9.9\%)$ and CSS(\texttt{RoBERTa}) by $(+3.1\%)$, even surpassing supervised RoBERTa(\texttt{DPR-ft}) by $(+1.4\%)$.
\\
\noindent\textbf{\emph{KnowRef}}~\cite{emami2019knowref}: a co-reference corpus addressing gender and number bias. The proposed approach outperforms the unsupervised baseline CSS(\texttt{BERT}) by a margin of $(+4.4\%)$ and CSS (\texttt{RoBERTa}) by $(+1.9\%)$.
\\
\noindent\textbf{\emph{WinoGender}}~\cite{rudinger-etal-2018-gender}: a gender-balanced co-reference corpus. The proposed approach outperforms the unsupervised baseline CSS(\texttt{BERT}) by a margin of $(+23.8\%)$ and CSS(\texttt{RoBERTa}) by $(+17.1\%)$.

\subsubsection{Attention-level Analysis}
Inspired by ~\cite{vig-belinkov-2019-analyzing}, we assess the impact of the attention mechanism by analyzing the attention tensor which is obtained by querying the attention of the \emph{MASK} token w.r.t. right/wrong candidate over all layers and heads.
The tensor decomposes into elements ${a}^{h,l}_{i,j}(x)$, gauging the influence of token $i$ w.r.t. token $j$ in layer $l$ of attention head $h$.
Aggregating the attention of \emph{MASK} token $i$ for the tokens $c_j$ for the right and wrong candidates by summation, slices the tensor into matrices $A_r, A_w\in \mathbb{R}^{H\times L}$ generating attention maps.Here $A_r, A_w$ corresponds to the attention maps w.r.t. the \emph{\underline{r}ight} answer and the \emph{\underline{w}rong} answer, respectively. Following~\cite{Brunner2020On}, we also investigated the maps of the \emph{last} $k$-layers, denoted as $A^{[k]}_r$ and $A^{[k]}_w$. We then computed the attention difference and entropy $H(.)$ difference on the attention maps of all DPR~\cite{rahman-ng-2012-resolving} samples, and presented the statistics in Tab.~\ref{tab:attention-stats}.

\begin{table}
\begin{tabular}{l|lll}
 & RoBa & CSS & \textbf{Ours} \\
 \hline\hline
$|H(A_r)-H(A_w)|$ & 0.024 & $\mathbf{0.097}$ & 0.078 \\
\hdashline
$|H(A^{[3]}_r)-H(A^{[3]}_w)|$ & 0.005 & 0.772 & $\textbf{1.328}$ \\
\hline
$|\bar{A}_r-\bar{A}_w|$ & 0.009 & 0.010 & $\mathbf{0.061}$ \\
\hdashline
$|\bar{A}^{[3]}_r-\bar{A}^{[3]}_w|$ & 0.020 & 0.034 & $\mathbf{0.306}$\\
\hline
\end{tabular}
\caption{Attention analysis of different models on DPR, and $k=3$. Top: entropies, Bottom: mean statistics.}
\label{tab:attention-stats}
\end{table}

We observed a significant concentration of attention for the right candidates for the proposed approach compared to the wrong ones. This pattern is even more pronounced for the last 3 layers. Specifically, we observed the manifestation of an average entropy of $3.41$ \emph{(right)} nats vs. $2.1$ nats \emph{(wrong)} on the last 3 layers, giving rise to the emergence of the desired pattern of more concerted attention on the right candidate. See supplementary material for more detailed results.

\subsubsection{Ablation Study}
To assess the contribution of each component, we evaluated the performance of each module
separately, gradually adding components to the loss.  See Tab.~\ref{tab:ablation-study} for the ablation study on WSC and WinoGrande. Pre-trained RoBERTa (\emph{large}) constitutes the baseline. \texttt{MEx} denotes the mutual-exclusive loss on the sentence log-likelihoods~\cite{klein-nabi-2020-contrastive}, \texttt{CA} denotes the contrastive attention defined in Sec.~\ref{sec:CA}, \texttt{CM} denotes the contrastive-margin defined in Sec.~\ref{sec:CM}.
While the CA term alone already suggests strong performance, this does not apply to the CM term. Given the regulatory nature of the CM term, optimizing it in isolation yields a model with inferior accuracy.

\begin{table}
\centering
\begin{tabular}{l|ll}
\textbf{Method} & \textbf{WSC} & \textbf{W.G.}\\
\hline
\hline
\centering
RoBERTa~\cite{liu2019roberta} &67.76 & 53.75 \\
CSS (\texttt{RoBERTa})& 79.85 & 57.77 \\
\hdashline
Our Method (\texttt{CM}) & 60.81 & 52.88 \\
Our Method (\texttt{CA}) & 80.95 & 57.14 \\
\textbf{Our Method} (\texttt{CA+CM}) & \textbf{84.10} & \textbf{60.80} \\
\hline
\end{tabular}
\caption{Ablation study, performance in accuracy (\%)}
\label{tab:ablation-study}
\end{table}

\section{Conclusion}
In this paper, we introduce an attention-level self-supervised learning method for commonsense reasoning. Specifically, we propose a method that enforces a contrastive loss on the attentions produced by transformer LM while pushing the likelihood of the candidates towards the extremities. The experimental analysis demonstrates that our proposed system outperforms the previous unsupervised state-of-the-art in multiple datasets.

\bibliography{anthology,custom}
\bibliographystyle{acl_natbib}

\end{document}